\documentclass[10pt,twocolumn,letterpaper]{article}

\usepackage{cvpr}
\usepackage{times}
\usepackage{epsfig}
\usepackage{graphicx}
\usepackage{amsmath}
\usepackage{amssymb}
\usepackage{amsfonts}

\usepackage{booktabs}
\usepackage{adjustbox}
\usepackage{array}
\usepackage{placeins}
\usepackage{afterpage}
\usepackage{stfloats}
\usepackage{textcomp}

\usepackage[pagebackref=true,breaklinks=true,letterpaper=true,colorlinks,bookmarks=false]{hyperref}

\cvprfinalcopy 

\def\cvprPaperID{2361} 

\newcommand{\secref}[1]{Section~\ref{sec:#1}}
\newcommand{\figref}[1]{Figure~\ref{fig:#1}}
\newcommand{\tabref}[1]{Table~\ref{tab:#1}}
\renewcommand{\etal}{~\emph{et al.}}

\newcolumntype{R}[2]{%
    >{\adjustbox{angle=#1,lap=\width-(#2)}\bgroup}%
    l%
    <{\egroup}%
}
\newcommand*\rot{\multicolumn{1}{R{45}{1em}}}

\newcommand{\loss}[1]{\mathcal{L}_{#1}}
\newcommand{\degrees}{$^{\circ}$ }

\ifcvprfinal\fi
\begin{document}

\title{Predicting Semantic Map Representations from Images \\using Pyramid Occupancy Networks}

\author{Thomas Roddick\\
University of Cambridge\\
{\tt\small tr346@cam.ac.uk}
\and
Roberto Cipolla\\
University of Cambridge\\
{\tt\small rc10001@cam.ac.uk}
}

\maketitle

\begin{abstract}
Autonomous vehicles commonly rely on highly detailed birds-eye-view maps of their environment, which capture both static elements of the scene such as road layout as well as dynamic elements such as other cars and pedestrians. Generating these map representations on the fly is a complex multi-stage process which incorporates many important vision-based elements, including ground plane estimation, road segmentation and 3D object detection. In this work we present a simple, unified approach for estimating maps directly from monocular images using a single end-to-end deep learning architecture.
For the maps themselves we adopt a semantic Bayesian occupancy grid framework, allowing us to trivially accumulate information over multiple cameras and timesteps.
We demonstrate the effectiveness of our approach by evaluating against several challenging baselines on the NuScenes and Argoverse datasets, and show that we are able to achieve a relative improvement of 9.1\% and 22.3\% respectively compared to the best-performing existing method.
\footnote{Source code and dataset splits will be made available at \url{github.com/tom-roddick/mono-semantic-maps}.}
\end{abstract}


\section{Introduction}

\begin{figure}
    \centering
    \begin{tabular}{c}
        \includegraphics[width=\linewidth]{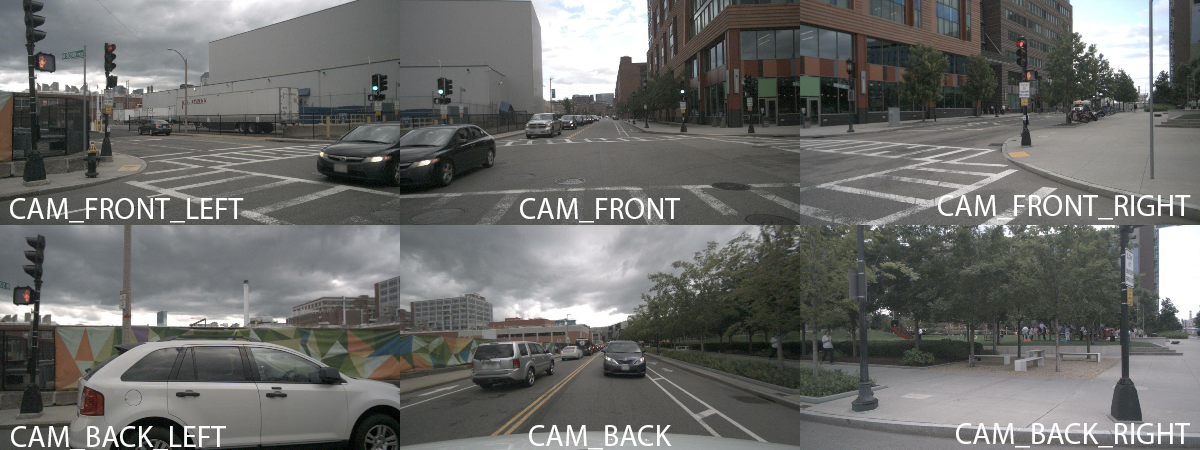} \\
        
        \includegraphics[width=\linewidth]{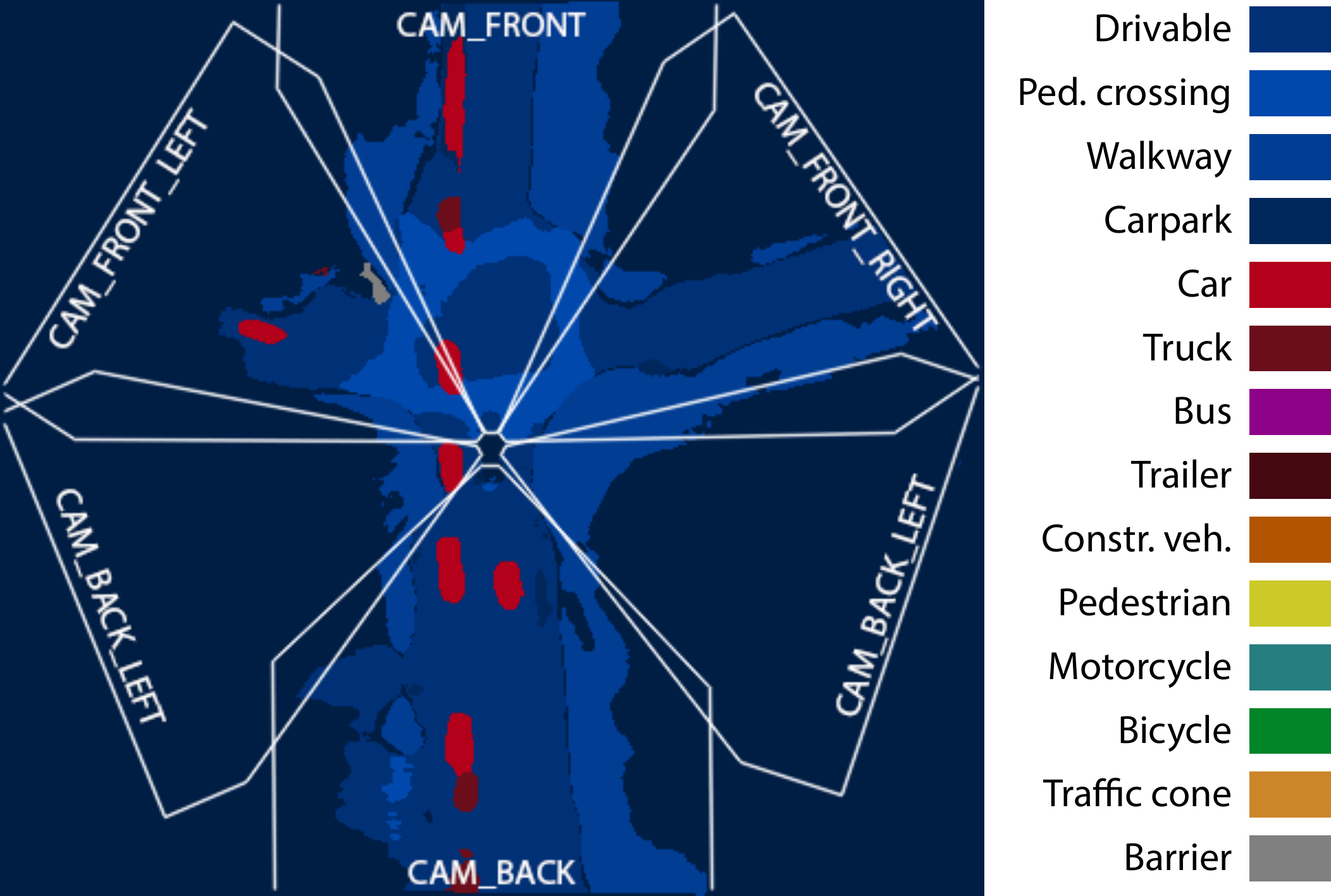} 
    \end{tabular}
    \caption{An example prediction from our algorithm. Given a set of surround-view images, we predict a full 360\degrees birds-eye-view semantic map, which captures both static elements like road and sidewalk as well as dynamic actors such as cars and pedestrians.}
    \label{fig:headline}
\end{figure}

Autonomous vehicles and other robotic platforms require a rich, succinct and detailed representation of their environment which captures both the geometry and layout of the static world as well as the pose and dimensions of other dynamic agents. Such representations often provide the foundation for all decision making, including path planning, collision avoidance and navigation. Rather than capturing the full 3D world in its entirety, one popular solution is to represent the world in the form of a birds-eye-view (BEV) map, which provide a compact way to capture the spatial configuration of the scene. Such maps are convenient in that they are simple to visualise and process, exploiting the fact that in many scenarios the essential information for navigation is largely confined to the ground plane. 

Construction of birds-eye-view maps is however at present a complex multistage processing pipeline, involving the composition of multiple fundamental machine vision tasks: structure from motion, ground plane estimation, road segmentation, lane detection, 3D object detection, and many more.
Intuitively, all these tasks are related: knowing the layout of the road ought to inform us about where in the image we should look for cars; and similarly a car emerging from behind a building may indicate the presence of a hidden side road beyond.  
There seems to be a clear impetus towards replacing this complicated pipeline with a simple end-to-end approach which is able to reason holistically about the world and predict the desired map representation directly from sensor observations.
In this work we focus on the particularly challenging scenario of BEV map estimation from monocular images alone. Given the high cost and limited resolution of LiDAR and radar sensors, the ability to build maps from image sensors alone is likely to be crucial to the development of robust autonomous vehicles.


Whilst a number of map representations are possible, we choose to represent the world using a probabilistic occupancy grid framework.
Occupancy grid maps~\cite{elfes1990occupancy} are widely used in robotics, and allow us to trivially incorporate information over multiple sensors and timesteps. Unlike other map representations, their grid-based structure also makes them highly agreeable to processing by convolutional neural networks, allowing us to take advantage of powerful developments from the deep learning literature.
In this work we extend the traditional definition of occupancy grids to that of a semantic occupancy grid~\cite{lu2019monocular}, which encodes the presence or absence of an object category at each grid location. Our objective is then to predict the probability that each semantic class is present at each location in our birds-eye-view map.

The contributions of this paper are as follows:
\begin{enumerate}
    \item We propose a novel dense transformer layer which maps image-based feature maps into the birds-eye-view space. 
    \item We design a deep convolutional neural network architecture, which includes a pyramid of transformers operating at multiple image scales, to predict accurate birds-eye-view maps from monocular images. 
    \item We evaluate our approach on two large-scale autonomous driving datasets, and show that we are able to considerably improve upon the performance of leading works in the literature.
\end{enumerate}
We also qualitatively demonstrate how a Bayesian semantic occupancy grid framework can be used to accumulate map predictions across multiple cameras and timesteps to build a complete model of a scene. The method is fast enough to be used in real time applications, processing 23.2 frames per second on a single GeForce RTX 2080 Ti graphics card.

\section{Related Work}
\paragraph{Map representations for autonomous driving}
High definition birds-eye-view maps have been shown to be an extremely powerful representation across a range of different driving tasks.  In 3D object detection, \cite{yang2018hdnet} use ground height prior information from maps to improve the quality of input LiDAR point clouds. \cite{ma2019exploiting} correlate visual observations with sparse HD map features to perform highly accurate localisation. Birds-eye-view maps are particularly valuable in the context of prediction and planning given their metric nature: \cite{djuric2018motion} and \cite{casas2018intentnet} render the local environment as a rasterised top-view map representation, incorporating road geometry, lane direction, and traffic agents, and use this representation to predict future vehicle trajectories. A similar representation is used by \cite{bansal2018chauffeurnet} as input to their imitation learning pipeline, allowing an autonomous agent to drive itself by recursively predicting its future state. \cite{hecker2018end} augment their camera-based end-to-end driving model with a rendered map view from a commercial GPS route planner and show that this significantly improves driving performance.      


\paragraph{Top-down representations from images}
A number of prior works have tackled the difficult problem of predicting birds-eye-view representations directly from monocular images. A common approach is to use inverse perspective mapping (IPM) to map front-view image onto the ground plane via a homography~\cite{ammar2019geometric, lin2012vision}. \cite{zhu2018generative} use a GAN to refine the resulting predictions. Other works focus on the birds-eye-view object detection task, learning a to map 2D bounding box detections to the top-down view\cite{palazzi2017learning, wang2019monocular}, or predicting 3D bounding boxes directly in the birds-eye-view space~\cite{roddick2019orthographic}.

Relatively few works however have tackled the more specific problem of generating semantic maps from images. Some use the IPM approach mentioned above to map a semantic segmentation of the image plane into the birds-eye-view space~\cite{deng2019restricted, samann2018efficient}, an approach which works well for estimating local road layout but which fails for objects such as cars and pedestrians which lie above the ground plane. \cite{henriques2018mapnet} take advantage of RGB-D images to learn an implicit map representation which can be used for later localisation. The VED method of \cite{lu2019monocular} uses a variational encoder-decoder network to predict a semantic occupancy grid directly from an image. The use of a fully-connected bottleneck layer in the network however means that much of the spatial context in the network is lost, leading to an output which is fairly coarse and is unable to capture small objects such as pedestrians. \cite{pan2019cross} adopt a similar approach, predicting a birds-eye-view semantic segmentation from a stack of surround view images, via a fully-connected view-transformer module. \cite{schulter2018learning} propose to use an in-painting CNN to infer the semantic labels and depth of the scene behind foreground objects, and generate a birds-eye-view by projecting the resulting semantic point cloud onto the ground plane. 

Unfortunately, given the lack of available ground truth data, many of the above methods are forced to rely on weak supervision from stereo~\cite{lu2019monocular}, weakly-aligned map labels~\cite{schulter2018learning} or synth-to-real domain transfer~\cite{schulter2018learning, pan2019cross}. Training on real data is crucial to performance in safety critical systems, and we believe we are the first to do so using a directly supervised approach.  

\begin{figure*}[t]
    \centering
    \includegraphics[width=\linewidth]{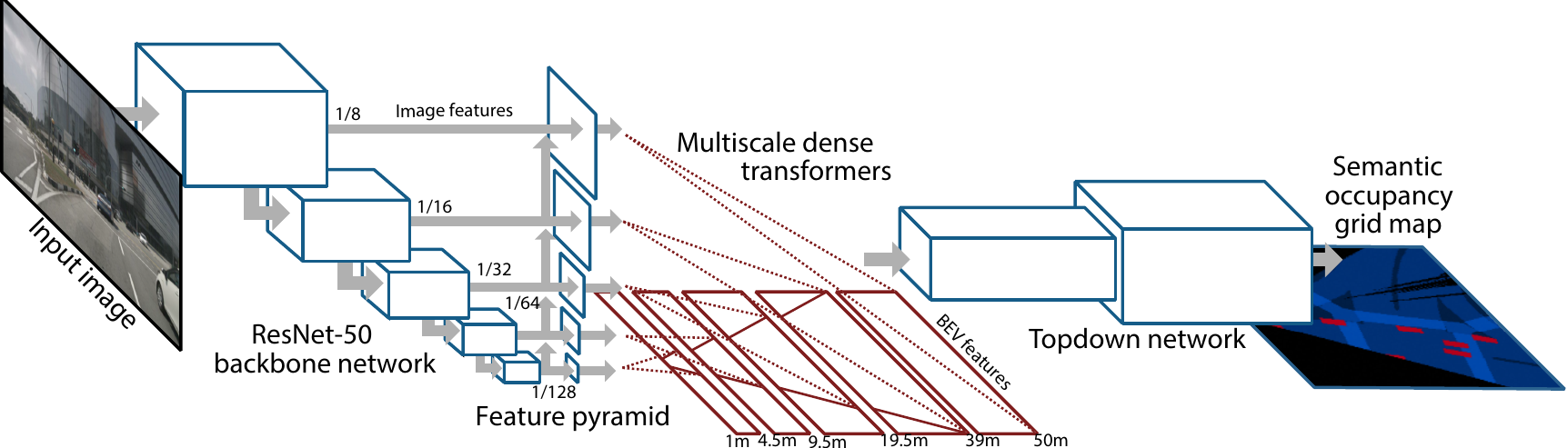}
    \caption{Architecture diagram showing an overview of our approach. (1) A ResNet-50 backbone network extracts image features at multiple resolutions. (2) A feature pyramid augments the high-resolution features with spatial context from lower pyramid layers. (3) A stack of dense transformer layers map the image-based features into the birds-eye-view. (4) The topdown network processes the birds-eye-view features and predicts the final semantic occupancy probabilities.}
    \label{fig:arch}
\end{figure*}

\section{Semantic occupancy grid prediction}

In this work we represent the state of the world as a birds-eye-view semantic occupancy grid map. Occupancy grid maps~\cite{elfes1990occupancy} are a type of discrete random field where each spatial location $x_i$ has an associated state $m_i$, which may be either occupied ($m_i=1$), or free ($m_i=0$). In practice, the true state of the world is unknown, so we treat $m_i$ as a random variable, and estimate the probability of occupancy $p(m_i|z_{1:t})$, conditioned on a set of observations $z_t$. The occupancy grid formulation may be further  extended to that of a semantic occupancy grid, where instead of generic cell occupancy, the state $m_i^c$ represents the presence or absence of an object of class $c$ in a given grid cell. These occupancies are non-exclusive: for example road, crossing and vehicle classes may conceivably coexist at the same location. 

Traditionally in occupancy grid mapping, the occupancy probabilities $p(m_i|z_{t}$) are estimated using an inverse sensor model, often a simple hand-engineered function which maps from range sensor readings to occupancy probabilities based on sensor characteristics. In our application, observations take the form of images and cell occupancies capture high-level semantic knowledge of the scene. We therefore propose to train a deep CNN-based inverse sensor model $p(m_i^c|z_t) = f_\theta(z_t, x_i)$ which learns to predict occupancy probabilities from a single monocular input image. 

Our objective is therefore to predict a set of multiclass binary labels at each location on a 2D birds-eye-view image. This scenario bears many similarities to the widely-studied computer vision problem of semantic segmentation. What makes this task particularly challenging however is the fact that the input and output representations exist within entirely different coordinate systems: the former in the perspective image space, and the latter in the orthographic birds-eye-view space. We therefore propose a simple transformer layer, which makes use of both camera geometry and fully-connected reasoning to map features from the image to the birds-eye-view space.

We incorporate this dense transformer layer as part of our deep Pyramid Occupancy Network (PyrOccNet). The pyramid occupancy network consists of four main stages. A \textbf{backbone feature extractor} generates multiscale semantic and geometric features from the image. This is then passed to an FPN~\cite{lin2017feature}-inspired \textbf{feature pyramid} which upsamples low-resolution feature-maps to provide context to features at higher resolutions. A stack of \textbf{dense transformer layers} together map the image-based features into the birds-eye view, which are processed by the \textbf{topdown network} to predict the final semantic occupancy grid probabilities. An overview of the approach is shown in \figref{arch}.


\FloatBarrier

\subsection{Losses}
\label{sec:loss}
We train our network using a combination of two loss functions. The binary cross entropy loss encourages the predicted semantic occupancy probabilities $p(m_i^c|z_t)$ to match the ground truth occupancies $\hat{m}_i^c$. Given that our datasets includes many small objects such as pedestrians, cyclists and traffic cones, we make use of a balanced variant of this loss, which up-weights occupied cells belonging to class $c$ by a constant factor $\alpha^c$:
\begin{equation}
\loss{xent} = \alpha^c \hat{m}_i^c \log p(m_i^c|z_t) + (1-\alpha^c)(1-\hat{m}_i^c)\log\left(1-p(m_i^c|z_t)\right)
\end{equation}

Neural networks are however renowned for routinely predicting high probabilities even in situations where they are highly uncertain. To encourage the networks to predict high uncertainty in regions which are known to be ambiguous, we introduce a second loss, which maximises the entropy of the predictions, encouraging them to fall close to 0.5:
\begin{equation}
\loss{uncert} = 1 - p(m_i^c|z_t)\log_2 p(m_i^c|z_t)
\end{equation}
We apply this maximum entropy loss only to grid cells which are not visible to the network, either because they fall outside field of view of the image, or because they are completely occluded (see \secref{data} for details). We ignore the cross entropy loss in these regions. The overall loss is given by the sum of the two loss functions:
\begin{equation}
\loss{total} = \loss{xent} + \lambda\loss{uncert}
\end{equation}
where $\lambda=0.001$ is a constant weighting factor.

\subsection{Temporal and sensor data fusion}
\label{sec:fusion}
The Bayesian occupancy grid formulation provides a natural way of combining information over multiple observations and multiple timesteps using a Bayesian filtering approach~\cite{thrun2005probabilistic}. Consider an image observation $z_t$ taken by a camera with extrinsic matrix $M_t$. We begin by converting our occupancy probabilities $p(m_i^c|z_t)$ into a log-odds representation
\begin{equation}
l_{i, t}^c = \log\frac{p(m_i^c|z_t)}{1-p(m_i^c|z_t)}
\end{equation}

which conveniently is equivalent to the network's pre-sigmoid output activations. The combined log-odds occupancies over observations $1$ to $t$ is then given by
\begin{equation}
l_{i, 1:t}^c = l_{i, 1:t-1}^c + l_{i, t}^c - l_0^c
\end{equation}
from which the occupancy probabilities after fusion can be recovered by applying the standard sigmoid function
\begin{equation}
    p(m_i^c|z_{1:t}) = \frac{1}{1 + \exp\left(-l_{i, 1:t}^c\right)}
\end{equation}
The log-odds value $l_0^c$ represents the prior probability of occupancy for class $c$:
\begin{equation}
l_0^c = \frac{p(m_i^c)}{1-p(m_i^c)}
\end{equation}

To obtain the occupancy probabilities in the global coordinate system, we resample the output from our network, which predicts occupancies in the local camera-frame coordinate system, into the global frame using the extrinsics matrix $M_t$, i.e. $p(m_i|z_t) = f_\theta(z_t, M_t^{-1}x_i)$. This approach is used in \secref{fusion-results} both to combine sensory information from a set of surround view cameras, and also to fuse occupancy grids over a 20s duration sequence of observations.

\subsection{Dense transformer layer}
\label{sec:transform}
\begin{figure}[t]
    \centering
    \includegraphics[width=\linewidth]{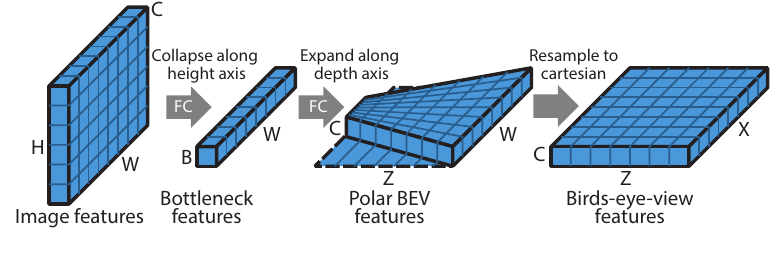}
    \caption{Our dense transformer layer first condenses the image-based features along the vertical dimension, whilst retaining the horizontal dimension. We then predict a set of features along the depth axis in a polar coordinate system, which are then resampled to Cartesian coordinates.}
    \label{fig:transform}
\end{figure}

One of the fundamental challenges of the occupancy grid prediction task is that the input and output exist in two entirely disparate coordinate systems: the perspective image space and the orthographic birds-eye-view space. To overcome this problem, we introduce a simple transformation layer, which is depicted in \figref{transform}. Our objective is to convert from an image plane feature map with $C$ channels, height $H$ and width $W$, to a feature map on the birds-eye-view plane with $C$ channels, depth $Z$ and width $X$. 

The dense transformer layer is inspired by the observation that while the network needs a lot of vertical context to map features to the birds-eye-view (due to occlusion, lack of depth information, and the unknown ground topology), in the horizontal direction the relationship between BEV locations and image locations can be established using simple camera geometry. Therefore, in order to retain the maximum amount of spatial information, we collapse the vertical dimension and channel dimensions of the image feature map to a bottleneck of size $B$, but preserve the horizontal dimension $W$. We then apply a 1D convolution along the horizontal axis and reshape the resulting feature map to give a tensor of dimensions $C\times Z\times W$. However this feature map, which is still in image-space coordinates, actually corresponds to a trapezoid in the orthographic birds-eye-view space due to perspective, and so the final step is to resample into a Cartesian frame using the known camera focal length $f$ and horizontal offset $u_0$.

\subsection{Multiscale transformer pyramid} 
\label{sec:multiscale}

The resampling step described in \secref{transform} involves, for a row of grid cells a distance $z$ away from the camera, sampling the polar feature map at intervals of 
\begin{equation}
    \Delta u = \frac{f\Delta x}{sz}
\end{equation}
where $\Delta x$ is the grid resolution and $s$ is the downsampling factor of input feature map with respect to the image. The use of a constant factor for $s$ however is problematic: features corresponding to grid cells far from the camera will be blurred whilst those close to the camera will be undersampled and aliasing can occur. We therefore propose to apply multiple transformers, acting on a pyramid of feature maps with downsampling factors $s_k = 2^{k+3}, k\in{\{0, ..., 4\}}$. The $k^{th}$ transformer generates features for a subset of depth values, ranging from $z_k$ to $z_{k-1}$, where $z_k$ is given by

\begin{equation}
    z_k = \frac{f\Delta x}{s_k}.
\end{equation}
Values of $z_k$ for a typical camera and grid setting are given in \tabref{depth}. The final birds-eye-view feature map is then constructed by concatenating the outputs of each individual transformer along the depth axis.

One downside of this approach is that at high resolutions the height of the feature maps $H_k$ can become very large, which leads to an excessive number of parameters in the corresponding dense transformer layer. In practice however, we can crop the feature maps to a height 
\begin{equation}
    H_k = f\frac{y_{max} - y_{min}}{s_kz_k}
\end{equation}
corresponding to a fixed vertical range between $y_{min}$ and $y_{max}$ in the world space. This means that the heights of the cropped feature maps stay roughly constant across scales.  

The feature maps are taken from the outputs of each residual stage in our backbone network, from conv3 to conv7. To ensure that the high resolution feature maps still encompass a large spatial context, we add upsampling layers from lower resolutions in the style of~\cite{lin2017feature}.

\begin{table}[tb]
\centering
\caption{Depth intervals for each layer of the feature pyramid.}
\begin{tabular}{@{}cccccc@{}}
\toprule
$k$           & 0     & 1     & 2     & 3     & 4     \\ \midrule
$s_k$           & 8     & 16    & 32    & 64    & 128   \\
$z_{k}$ (m)   & 39.0   & 19.5   & 9.0   & 4.5  & 1.0  \\
ResNet layer     & conv3 & conv4 & conv5 & conv6 & conv7 \\ \bottomrule
\end{tabular}
\label{tab:depth}
\end{table}

\section{Experimental Setup}

\subsection{Datasets}
We evaluate our approach against two large-scale autonomous driving datasets. The NuScenes dataset~\cite{caesar2019nuscenes} consists of 1000 short video sequences captured from four locations in Boston and Singapore. It includes images captured from six calibrated surround-view cameras, 3D bounding box annotations for 23 object categories and rich semantic map annotations which include vectorised representations of lanes, traffic lights, sidewalks and more. From these we select a subset of four map categories which can feasibly be estimated from images, along with ten object categories. 

The Argoverse 3D dataset~\cite{chang2019argoverse} is comprised of 65 training and 24 validation sequences captured in two cities, Miami and Pittsburg, using a range of sensors including seven surround-view cameras. Like NuScenes, the Argoverse dataset provides both 3D object annotations from 15 object categories, as well as semantic map information including road mask, lane geometry and ground height. From these we choose 7 object categories which contain sufficient training examples, along with the driveable road mask.

As both NuScenes and Argoverse are predominantly object detection rather than map prediction datasets, the default dataset splits contain multiple road segments which appear in both the training and validation splits. We therefore redistribute the train/val sequences to remove any overlapping segments, taking care to ensure a balanced distribution over locations, objects and weather conditions.

\begin{table*}[t!]
\centering
\caption{Intersection over Union scores (\%) on the Argoverse dataset. CS Mean is the average over classes present in the Cityscapes dataset, indicated by *. Letters beside the method represent the presence of each component in our ablation study: D - dense transformer layer, P - transformer pyramid, T - topdown network.}
\label{tab:argoverse}
\begin{tabular}{@{}c|cccccccc|cc@{}}
\toprule
Method & Drivable* & Vehicle* & Pedest.* & Large veh. & Bicycle* & Bus* & Trailer & Motorcy.* & Mean & CS Mean \\
\midrule
IPM & 43.7 & 7.5 & 1.5 & - & 0.4 & 7.4 & - & 0.8 & - & 10.2 \\
Depth Unproj. & 33.0 & 12.7 & 3.3 & - & 1.1 & \textbf{20.6} & - & 1.6 & - & 12.1 \\
VED \cite{lu2019monocular} & 62.9 & 14.0 & 1.0 & 3.9 & 0.0 & 12.3 & 1.3 & 0.0 & 11.9 & 15.0 \\
VPN \cite{pan2019cross} & 64.9 & 23.9 & 6.2 & 9.7 & 0.9 & 3.0 & 0.4 & 1.9 & 13.9 & 16.8 \\
\midrule
Ours - baseline & 58.5 & 23.4 & 3.9 & 5.2 & 0.5 & 11.0 & 0.4 & 1.9 & 13.1 & 16.5\\
Ours - D & 63.8 & 27.9 & 4.8 & 8.8 & 1.0 & 11.0 & 0.0 & 3.4 & 15.1 & 18.7 \\
Ours - D+P & \textbf{65.9} & 30.7 & 7.3 & 10.2 & 1.7 & 9.3 & \textbf{1.7} & 2.2 & 16.1 & 19.5\\
\midrule
Ours - D+P+T & 65.4 & \textbf{31.4} & \textbf{7.4} & \textbf{11.1} & \textbf{3.6} & 11.0 & 0.7 & \textbf{5.7} & \textbf{17.0} & \textbf{20.8} \\
\bottomrule
\end{tabular}
\end{table*}

\subsection{Data generation}
\label{sec:data}
The NuScenes and Argoverse datasets provide ground truth annotations in the form of vectorised city-level map labels and 3D object bounding boxes. We convert these into ground truth occupancy maps by first mapping all vector annotations into the coordinate system of the $t^{th}$ sample using the camera extrinsic matrix $M_t$ provided by the datasets. We then rasterise each annotation to a binary image in the birds-eye-view, which lies on a grid extending 50m in front of the given camera and 25m to either side, at a resolution of 25cm per pixel. For the case of object annotations, we first project the 3D bounding box onto the xz-plane to obtain a 2D polygon. The result of this process is a stack of binary images, which represent the ground truth occupancies for each semantic category $c$ as observed from camera $t$.

The resulting labels however represent a close to impossible task for the network, since some grid cell locations lie outside the camera field of view (FoV) or are completely occluded by other objects. We therefore generate an additional binary mask indicating whether each grid cell is visible. A cell is treated as visible if it is within the FoV and has at least one LiDAR ray passing through it (i.e. not blocked by a closer object). 


\subsection{Baselines}
\label{sec:baseline}

\paragraph{Published methods}
In order to demonstrate the effectiveness of our approach, we compare against two previously published works: the Variational Encoder-Decoder (VED) of Lu~\etal~\cite{lu2019monocular}, and the View Parsing Network (VPN) of Pan~\etal~\cite{pan2019cross}. These networks presume different input and output dimensions, so we make minor architectural changes which we detail in Section~A of the supplementary material. 

\paragraph{Inverse Perspective Mapping (IPM)}
We present a simple baseline inspired by other works~\cite{deng2019restricted, samann2018efficient} of mapping an image-based semantic segmentation to the ground plane via a homography. The image-level segmentation is computed using a state-of-the-art DeepLabv3~\cite{chen2017rethinking} network, pretrained on Cityscapes~\cite{cordts2016cityscapes}, which shares many classes in common with both NuScenes and Argoverse. The ground planes are obtained either by fitting a plane to LiDAR points in the case of NuScenes, or using the precomputed ground heights provided by Argoverse. Note that this information would not be available to a real monocular system at test time, making this baseline additionally competitive. 

\paragraph{Depth-based unprojection}
Another intuitive solution to this problem would be to use a monocular depth estimator to generate a 3D point cloud from the image, and then drop the z-axis to transfer image-based semantic labels onto the ground plane. As an upper-bound on the performance of this type of approach, we use ground truth depth computed by densifying LiDAR points using the algorithm adopted in the NYU depth dataset~\cite{silberman2012indoor, levin2004colorization}. We use the same DeepLabv3 to predict image level labels as before.

\subsection{Architecture and training details}
For the backbone and feature pyramid components of our network, we use a pretrained FPN network~\cite{lin2017feature}, which incorporates a ResNet-50~\cite{he2016deep} front-end. The topdown network consists of a stack of 8 residual blocks, including a transposed convolution layer which upsamples the birds-eye-view features from a resolution of 0.5m to 0.25m per pixel. For the balanced loss weighting $\alpha^c$, we use the square root of the inverse class frequency, as we found that using inverse frequency directly leads to a tendancy to overpredict on small classes. The uncertainty loss weighting $\lambda$ is taken as 0.001. We train all networks until convergence using SGD with a learning rate of 0.1, batch size 12 and a momentum of 0.9.


\subsection{Evaluation}
Our primary evaluation metric is the Intersection over Union (IoU) score, which we compute by binarising the predictions according to a Bayesian decision boundary ($p(m_i^c|z_t) > 0.5$). To account for the arbitrary nature of this threshold, we also provide precision-recall curves as part of the supplementary material. Non-visible grid cells (see \secref{data}) are ignored during evaluation.

\afterpage{%

\begin{figure*}[htp]
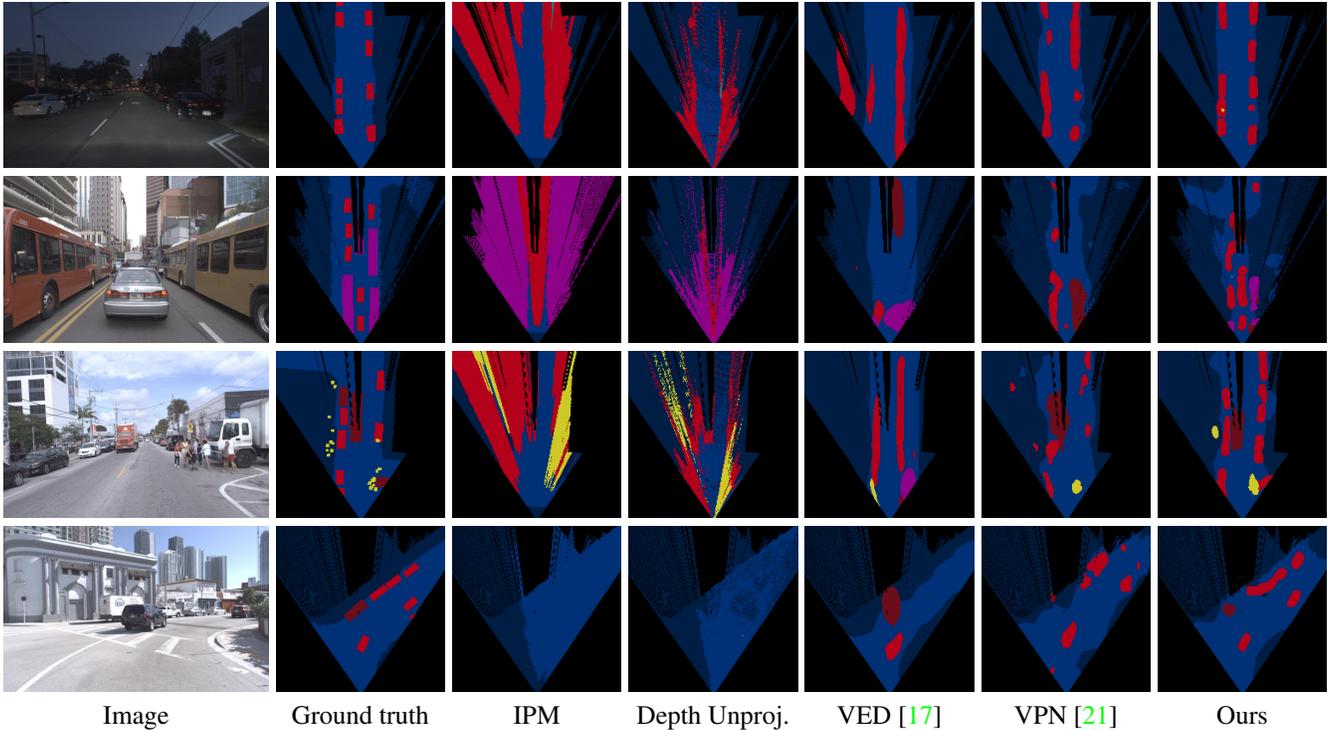

    \centering
    \setlength\tabcolsep{.05cm}
    \newcommand{\imggrid}[2][2.2cm]{\includegraphics[height=##1]{figures/argoverse/##2}}
    \begin{tabular}{ccccccc}
        
        \imggrid{0336_front_center_0206/image.jpg} &
        \imggrid{0336_front_center_0206/ground_truth.png} &
        \imggrid{0336_front_center_0206/plane.png} &
        \imggrid{0336_front_center_0206/depth.png} &
        \imggrid{0336_front_center_0206/vae.png} &
        \imggrid{0336_front_center_0206/vpn.png} &
        \imggrid{0336_front_center_0206/ours.png} \\
        
        \imggrid{3373_front_center_0037/image.jpg} &
        \imggrid{3373_front_center_0037/ground_truth.png} &
        \imggrid{3373_front_center_0037/plane.png} &
        \imggrid{3373_front_center_0037/depth.png} &
        \imggrid{3373_front_center_0037/vae.png} &
        \imggrid{3373_front_center_0037/vpn.png} &
        \imggrid{3373_front_center_0037/ours.png} \\
        
        \imggrid{6f15_front_center_0086/image.jpg} &
        \imggrid{6f15_front_center_0086/ground_truth.png} &
        \imggrid{6f15_front_center_0086/plane.png} &
        \imggrid{6f15_front_center_0086/depth.png} &
        \imggrid{6f15_front_center_0086/vae.png} &
        \imggrid{6f15_front_center_0086/vpn.png} &
        \imggrid{6f15_front_center_0086/ours.png} \\
        
        \imggrid{ba06_front_center_0035/image.jpg} &
        \imggrid{ba06_front_center_0035/ground_truth.png} &
        \imggrid{ba06_front_center_0035/plane.png} &
        \imggrid{ba06_front_center_0035/depth.png} &
        \imggrid{ba06_front_center_0035/vae.png} &
        \imggrid{ba06_front_center_0035/vpn.png} &
        \imggrid{ba06_front_center_0035/ours.png} \\

        Image & Ground truth & IPM & Depth Unproj. & VED \cite{lu2019monocular} & VPN~\cite{pan2019cross} & Ours \\ 
    \end{tabular}
    \caption{Qualitative results on the Argoverse dataset. For each grid location $i$, we visualise the class with the largest index $c$ which has occupancy probability $p(m_i^c|z_t) > 0.5$. Black regions (outside field of view or no lidar returns) are ignored during evaluation. See \figref{headline} for a complete class legend.}
    \label{fig:argoverse}
\end{figure*}

\begin{figure*}[htp]
    \centering
    \setlength\tabcolsep{.05cm}
    \newcommand{\imggrid}[2][2.2cm]{\includegraphics[height=##1]{figures/nuscenes/##2}}
    \begin{tabular}{ccccccc}
        \imggrid{0053_front_04_20_19.5/image.jpg} &
        \imggrid{0053_front_04_20_19.5/ground_truth.png} &
        \imggrid{0053_front_04_20_19.5/plane.png} &
        \imggrid{0053_front_04_20_19.5/depth.png} &
        \imggrid{0053_front_04_20_19.5/vae.png} &
        \imggrid{0053_front_04_20_19.5/vpn.png} &
        \imggrid{0053_front_04_20_19.5/ours.png} \\
        
        
        \imggrid{0391_front_16_59_05.1/image.jpg} &
        \imggrid{0391_front_16_59_05.1/ground_truth.png} &
        \imggrid{0391_front_16_59_05.1/plane.png} &
        \imggrid{0391_front_16_59_05.1/depth.png} &
        \imggrid{0391_front_16_59_05.1/vae.png} &
        \imggrid{0391_front_16_59_05.1/vpn.png} &
        \imggrid{0391_front_16_59_05.1/ours.png} \\
        
        \imggrid{0530_front_20_56_18.4/image.jpg} &
        \imggrid{0530_front_20_56_18.4/ground_truth.png} &
        \imggrid{0530_front_20_56_18.4/plane.png} &
        \imggrid{0530_front_20_56_18.4/depth.png} &
        \imggrid{0530_front_20_56_18.4/vae.png} &
        \imggrid{0530_front_20_56_18.4/vpn.png} &
        \imggrid{0530_front_20_56_18.4/ours.png} \\
        
        
        \imggrid{0747_front_20_20_17.7/image.jpg} &
        \imggrid{0747_front_20_20_17.7/ground_truth.png} &
        \imggrid{0747_front_20_20_17.7/plane.png} &
        \imggrid{0747_front_20_20_17.7/depth.png} &
        \imggrid{0747_front_20_20_17.7/vae.png} &
        \imggrid{0747_front_20_20_17.7/vpn.png} &
        \imggrid{0747_front_20_20_17.7/ours.png} \\

        Image & Ground truth & IPM & Depth Unproj. & VED \cite{lu2019monocular} & VPN~\cite{pan2019cross} & Ours \\ 
    \end{tabular}
    \caption{Qualitative results on the NuScenes dataset. See \figref{headline} for a complete class legend.}
    \label{fig:nuscenes}
    
\end{figure*} 

}

\FloatBarrier
\section{Results}

\subsection{Ablation study}

Before comparing against other methods, we validate our choice of architecture by performing an ablation study on the Argoverse dataset. We begin from a simple baseline, consisting of the backbone network, an inverse perspective mapping to geometrically map features to the birds-eye-view, and a sigmoid layer to predict final occupancy probabilities. We then incrementally reintroduce each of the key components of our approach: the dense transformer layer~(\textbf{D}), transformer pyramid~(\textbf{P}), and topdown network~(\textbf{T}).


The results of this ablation study are shown in the second half of \tabref{argoverse}. Each successive component improves the performance by a consistent factor of roughly 1\% mean IoU, with the addition of the dense transformer having a particularly pronounced effect on the results, which we argue is one of the key novelties of our approach. The topdown network provides no advantage for large classes such as driveable area, but significantly improves performance for small, rare classes such as motorbike and bicycle.


\begin{table*}[t!]
\centering
\caption{Intersection over Union scores (\%) on the NuScenes dataset. CS mean is the average over classes present in the Cityscapes dataset, which are indicated by *.}
\label{tab:nuscenes}
\begin{tabular}{@{}c|cccccccccccccc|cc@{}}
\toprule
Method & \rot{Drivable*} & \rot{Ped. crossing} & \rot{Walkway*} & \rot{Carpark} & \rot{Car*} & \rot{Truck} & \rot{Bus*} & \rot{Trailer} & \rot{Constr. veh.} & \rot{Pedestrian*} & \rot{Motorcycle*} & \rot{Bicycle*} & \rot{Traf. Cone} & \rot{Barrier} & \rot{Mean} & \rot{CS Mean} \\
\midrule
IPM & 40.1 & - & 14.0 & - & 4.9 & - & 3.0 & - & - &  0.6 & 0.8 & 0.2 & - & - & - & 9.1\\
Depth Unproj. & 27.1 & - & 14.1 & - & 11.3 & - & 6.7 & - & - & 2.2 & 2.8 & 1.3 & - & - & - & 9.4 \\
VED \cite{lu2019monocular} & 54.7 & 12.0 & 20.7 & 13.5 & 8.8 & 0.2 & 0.0 & 7.4 & 0.0 & 0.0 & 0.0 & 0.0 & 0.0 & 4.0 & 8.7 & 12.0 \\
VPN~\cite{pan2019cross} & 58.0 & 27.3 & 29.4 & 12.9 & \textbf{25.5} & \textbf{17.3} & 20.0 & \textbf{16.6} & 4.9 & 7.1 & 5.6 & 4.4 & 4.6 & \textbf{10.8} & 17.5 & 21.4 \\ \midrule
Ours & \textbf{60.4} & \textbf{28.0} & \textbf{31.0} & \textbf{18.4} & 24.7 & 16.8 & \textbf{20.8} & \textbf{16.6} & \textbf{12.3} & \textbf{8.2} & \textbf{7.0} & \textbf{9.4} & \textbf{5.7} & 8.1 & \textbf{19.1} & \textbf{23.1} \\
\bottomrule
\end{tabular}
\end{table*}



\subsection{Comparison to other methods}
\label{sec:argoverse}
In addition to the ablation experiments described above, we evaluate our final architecture to a number of baseline methods described in \secref{baseline}. It can be seen from \tabref{argoverse} that we outperform all previous approaches by a significant margin. The two prior works, VPN and VED, achieve a comparable IoU on the drivable area class (representing the road surface), but across the smaller classes such as vehicle, pedestrian etc., we are able to obtain considerably better results. We suggest that this improvement is explained by the fact that our dense transformer layer preserves more spatial information compared to the fully connected bottlenecks of~\cite{lu2019monocular} and \cite{pan2019cross}. This hypothesis is supported by the qualitative results illustrated in \figref{argoverse}, which show that our method is much more able to resolve fine details such as the separation between individual cars (rows 1 and 2) or crowds of pedestrians (row 3). Both VPN and in particular VED on the other hand are only capable of making relatively coarse predictions and often miss important features, such as the car in row 3. The IPM baseline achieves reasonably good performance on the drivable area class but fails across all other classes because the predictions are elongated along the camera rays, as can be seen from \figref{argoverse}. The success of the depth unprojection method meanwhile is limited by the inherent sparsity of the lidar point clouds beyond a range of about 25m.


\subsection{Evaluation on the NuScenes dataset}
\label{sec:nuscenes}

Having justified our approach on the relatively small Agoverse dataset, we move to the more challenging evaluation scenario of the NuScenes dataset. We report quantitative results in \tabref{nuscenes}, and visualise our predictions in \figref{nuscenes}. Despite the greater diversity of this dataset, we are able to outperform the next-best approach, the VPN method of \cite{pan2019cross}, by a relative factor of 9.1\%. As with Argoverse, our method is consistently able to capture finer details in the scene, such as the shape of the bus in row 2 and the geometry of the crossroads in row 3. On this dataset, the VED method completely breaks down for the cases of small (pedestrian, cyclist, etc.) or infrequently occurring (construction vehicle, bus) classes.

\subsection{Temporal and sensor fusion}
\label{sec:fusion-results}

Predicting BEV maps from a single viewpoint as discussed in \secref{nuscenes} and \secref{argoverse} is typically insufficient for driving purposes; in general we want to build a complete picture of our environment taking into account multiple sensors and historical information. In \figref{headline} we show an example of how the occupancy grids from six surround-view cameras can be combined using the Bayesian fusion scheme described in \secref{fusion}. We assume a prior probability of $p(m_i^c) = 0.5$ for all classes.

\begin{figure}[t!]
    \centering
    \setlength\tabcolsep{.1cm}
    \begin{tabular}{cc}
        \includegraphics[height=4.3cm]{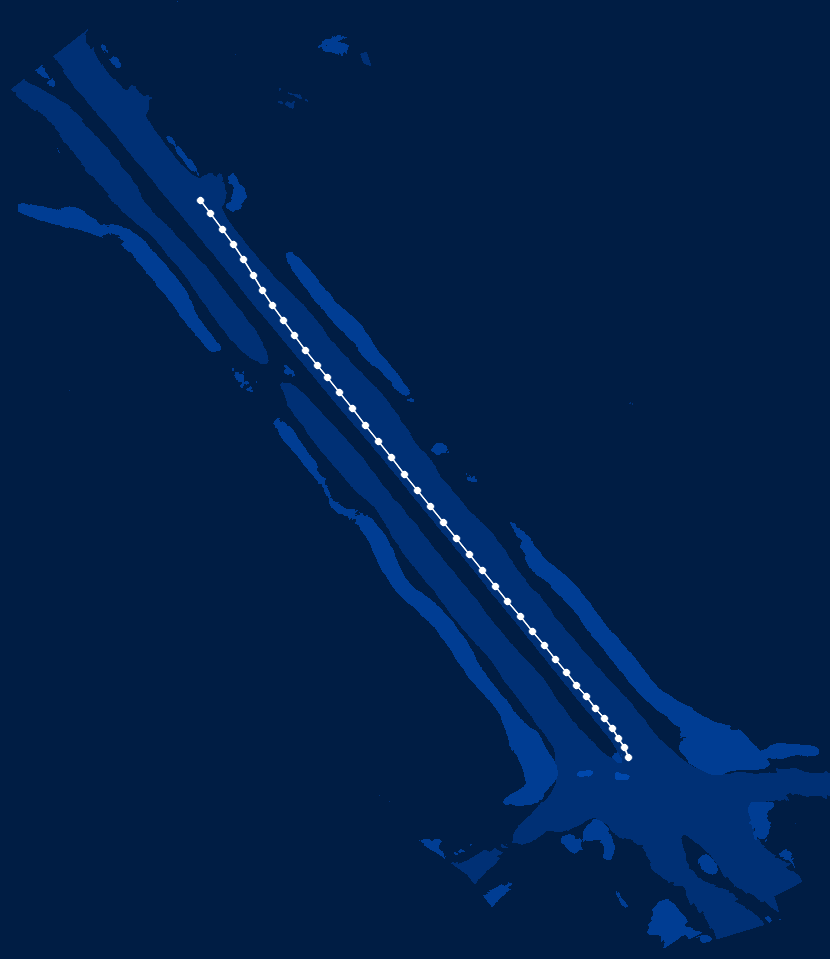} & 
        \includegraphics[height=4.3cm]{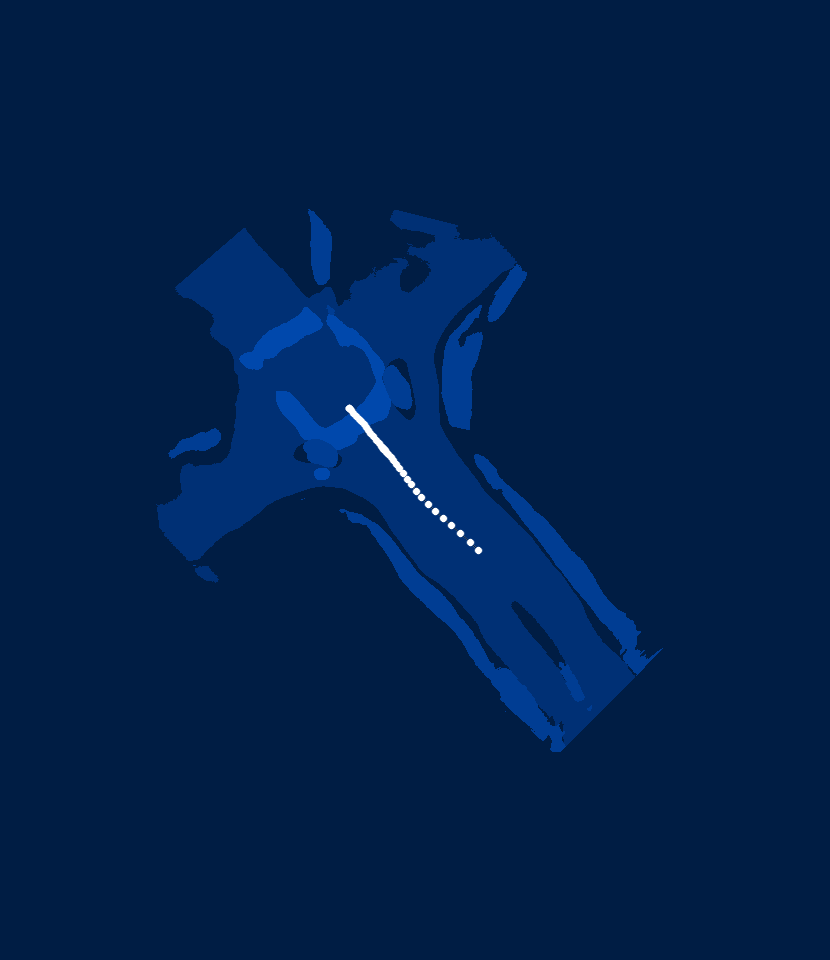} \\
        \includegraphics[height=4.3cm]{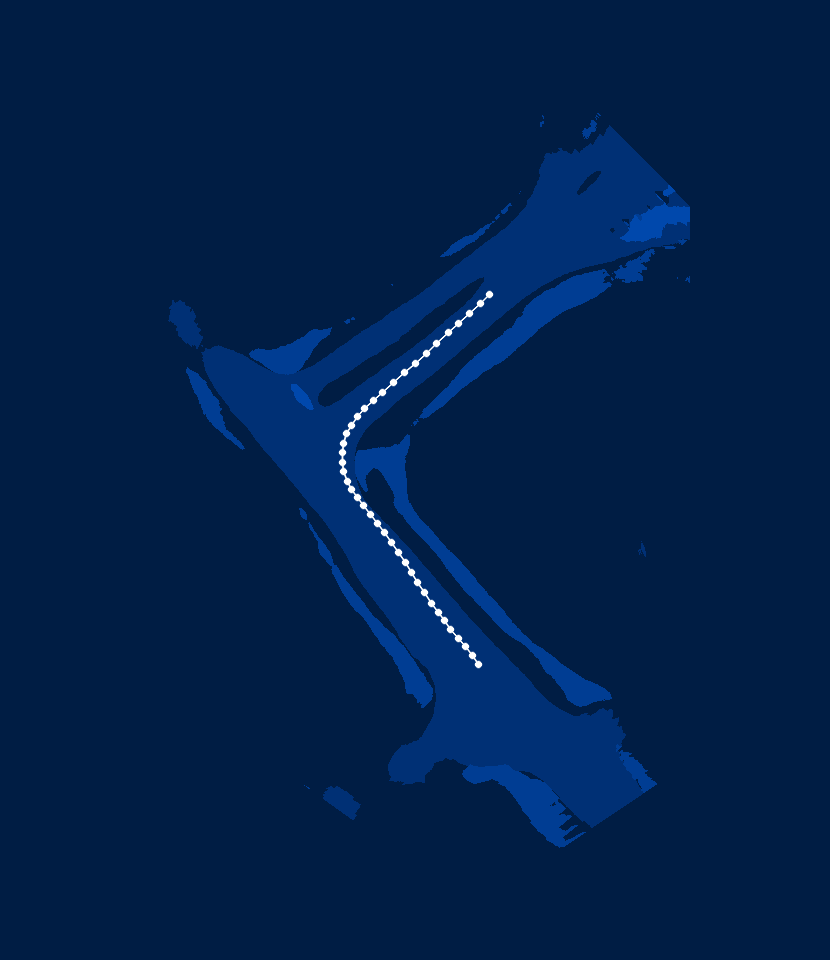} & 
        \includegraphics[height=4.3cm]{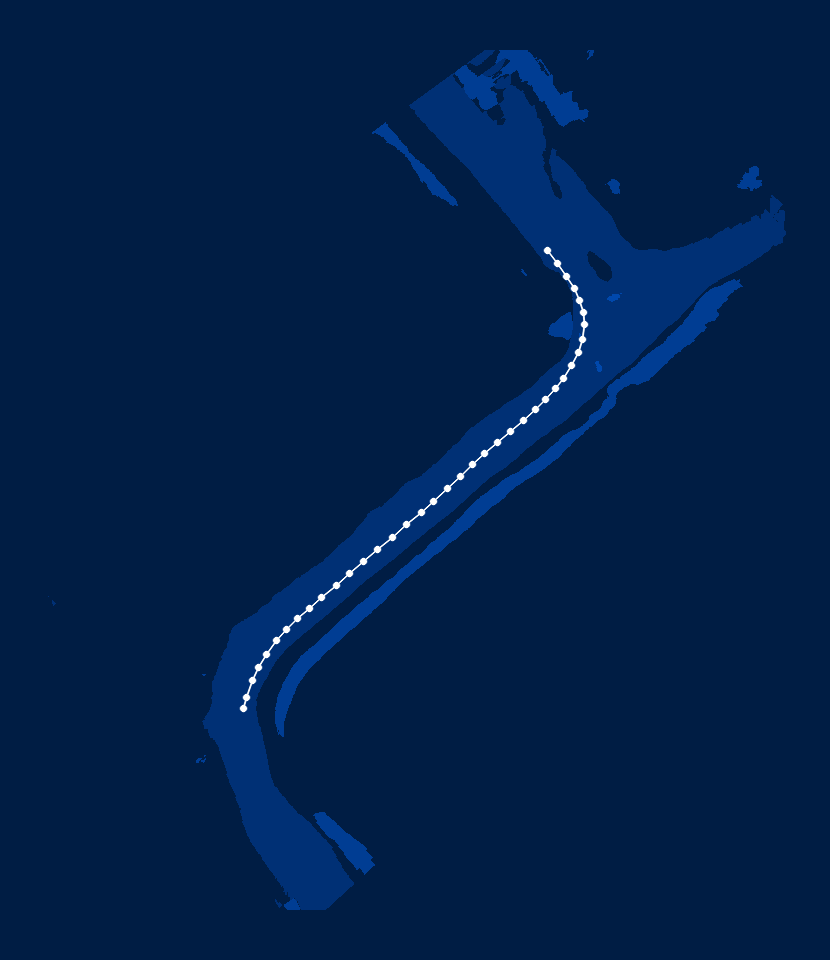}
    \end{tabular}
    \caption{Scene-level occupancy grid maps generated by accumulating occupancy probabilities over 20s sequences. White lines indicate the ego-vehicle trajectory. Note that only static classes (drivable, crossing, walkway, carpark) are visualised.}
    \label{fig:multiframe}
\end{figure}

For static elements of the scene, such as road, sidewalk etc., we can go a step further by combining predictions over multiple timesteps to build a complete model of the geometry of a given scene. \figref{multiframe} shows several examples of accumulating occupancy probabilities over 20s long sequences from the NuScenes dataset. The network is able to utilise information from multiple views to resolve ambiguities, resulting in a smoother overall prediction.



\section{Conclusions}
We have proposed a novel method for predicting birds-eye-view maps directly from monocular images. Our approach improves on the state-of-the-art by incorporating dense transformer layers, which make use of camera geometry to warp image-based features to the birds-eye-view, as part of a multiscale transformer pyramid. As well as predicting maps from a single image, our method is able to effortlessly combine information across multiple views to build an exhaustive model of the surrounding environment. We believe that this work provides a broad framework for future work into other tasks which operate in the birds-eye-view, such as lane instance detection and future prediction.  


\FloatBarrier
\clearpage

{\small
\bibliographystyle{ieee_fullname}
\bibliography{egbib}

\begin{thebibliography}{10}\itemsep=-1pt

\bibitem{ammar2019geometric}
Syed Ammar~Abbas and Andrew Zisserman.
\newblock A geometric approach to obtain a bird's eye view from an image.
\newblock {\em arXiv preprint arXiv:1905.02231}.

\bibitem{bansal2018chauffeurnet}
Mayank Bansal, Alex Krizhevsky, and Abhijit Ogale.
\newblock Chauffeurnet: Learning to drive by imitating the best and
  synthesizing the worst.
\newblock {\em arXiv preprint arXiv:1812.03079}, 2018.

\bibitem{caesar2019nuscenes}
Holger Caesar, Varun Bankiti, Alex~H Lang, Sourabh Vora, Venice~Erin Liong,
  Qiang Xu, Anush Krishnan, Yu Pan, Giancarlo Baldan, and Oscar Beijbom.
\newblock Nuscenes: A multimodal dataset for autonomous driving.
\newblock {\em arXiv preprint arXiv:1903.11027}, 2019.

\bibitem{casas2018intentnet}
Sergio Casas, Wenjie Luo, and Raquel Urtasun.
\newblock Intentnet: Learning to predict intention from raw sensor data.
\newblock In {\em Conference on Robot Learning}, pages 947--956, 2018.

\bibitem{chang2019argoverse}
Ming-Fang Chang, John Lambert, Patsorn Sangkloy, Jagjeet Singh, Slawomir Bak,
  Andrew Hartnett, De Wang, Peter Carr, Simon Lucey, Deva Ramanan, et~al.
\newblock Argoverse: 3d tracking and forecasting with rich maps.
\newblock In {\em Proceedings of the IEEE Conference on Computer Vision and
  Pattern Recognition (CVPR)}, pages 8748--8757, 2019.

\bibitem{chen2017rethinking}
Liang-Chieh Chen, George Papandreou, Florian Schroff, and Hartwig Adam.
\newblock Rethinking atrous convolution for semantic image segmentation.
\newblock {\em arXiv preprint arXiv:1706.05587}, 2017.

\bibitem{cordts2016cityscapes}
Marius Cordts, Mohamed Omran, Sebastian Ramos, Timo Rehfeld, Markus Enzweiler,
  Rodrigo Benenson, Uwe Franke, Stefan Roth, and Bernt Schiele.
\newblock The cityscapes dataset for semantic urban scene understanding.
\newblock In {\em Proceedings of the IEEE Conference on Computer Vision and
  Pattern Recognition (CVPR)}, 2016.

\bibitem{deng2019restricted}
Liuyuan Deng, Ming Yang, Hao Li, Tianyi Li, Bing Hu, and Chunxiang Wang.
\newblock Restricted deformable convolution-based road scene semantic
  segmentation using surround view cameras.
\newblock {\em IEEE Transactions on Intelligent Transportation Systems}, 2019.

\bibitem{djuric2018motion}
Nemanja Djuric, Vladan Radosavljevic, Henggang Cui, Thi Nguyen, Fang-Chieh
  Chou, Tsung-Han Lin, and Jeff Schneider.
\newblock Motion prediction of traffic actors for autonomous driving using deep
  convolutional networks.
\newblock {\em arXiv preprint arXiv:1808.05819}, 2018.

\bibitem{elfes1990occupancy}
Alberto Elfes et~al.
\newblock Occupancy grids: A stochastic spatial representation for active robot
  perception.
\newblock In {\em Proceedings of the Sixth Conference on Uncertainty in AI},
  volume 2929, page~6, 1990.

\bibitem{he2016deep}
Kaiming He, Xiangyu Zhang, Shaoqing Ren, and Jian Sun.
\newblock Deep residual learning for image recognition.
\newblock In {\em Proceedings of the IEEE Conference on Computer Vision and
  Pattern Recognition (CVPR)}, pages 770--778, 2016.

\bibitem{hecker2018end}
Simon Hecker, Dengxin Dai, and Luc Van~Gool.
\newblock End-to-end learning of driving models with surround-view cameras and
  route planners.
\newblock In {\em Proceedings of the European Conference on Computer Vision
  (ECCV)}, pages 435--453, 2018.

\bibitem{henriques2018mapnet}
Joao~F Henriques and Andrea Vedaldi.
\newblock Mapnet: An allocentric spatial memory for mapping environments.
\newblock In {\em proceedings of the IEEE Conference on Computer Vision and
  Pattern Recognition (CVPR)}, pages 8476--8484, 2018.

\bibitem{levin2004colorization}
Anat Levin, Dani Lischinski, and Yair Weiss.
\newblock Colorization using optimization.
\newblock In {\em ACM Transactions on Graphics (TOG)}, volume~23, pages
  689--694. ACM, 2004.

\bibitem{lin2012vision}
Chien-Chuan Lin and Ming-Shi Wang.
\newblock A vision based top-view transformation model for a vehicle parking
  assistant.
\newblock {\em Sensors}, 12(4):4431--4446, 2012.

\bibitem{lin2017feature}
Tsung-Yi Lin, Piotr Doll{\'a}r, Ross Girshick, Kaiming He, Bharath Hariharan,
  and Serge Belongie.
\newblock Feature pyramid networks for object detection.
\newblock In {\em Proceedings of the IEEE Conference on Computer Vision and
  Pattern Recognition (CVPR)}, pages 2117--2125, 2017.

\bibitem{lu2019monocular}
Chenyang Lu, Marinus Jacobus~Gerardus van~de Molengraft, and Gijs Dubbelman.
\newblock Monocular semantic occupancy grid mapping with convolutional
  variational encoder--decoder networks.
\newblock {\em IEEE Robotics and Automation Letters}, 4(2):445--452, 2019.

\bibitem{ma2019exploiting}
Wei-Chiu Ma, Ignacio Tartavull, Ioan~Andrei B{\^a}rsan, Shenlong Wang, Min Bai,
  Gellert Mattyus, Namdar Homayounfar, Shrinidhi~Kowshika Lakshmikanth, Andrei
  Pokrovsky, and Raquel Urtasun.
\newblock Exploiting sparse semantic {HD} maps for self-driving vehicle
  localization.
\newblock In {\em Proceedings of the IEEE/RSA International Conference on
  Intelligent Robots and Systems}, 2019.

\bibitem{silberman2012indoor}
Pushmeet~Kohli Nathan~Silberman, Derek~Hoiem and Rob Fergus.
\newblock Indoor segmentation and support inference from rgbd images.
\newblock In {\em Proceedings of the European Conference on Computer Vision
  (ECCV)}, 2012.

\bibitem{palazzi2017learning}
Andrea Palazzi, Guido Borghi, Davide Abati, Simone Calderara, and Rita
  Cucchiara.
\newblock Learning to map vehicles into bird’s eye view.
\newblock In {\em International Conference on Image Analysis and Processing},
  pages 233--243. Springer, 2017.

\bibitem{pan2019cross}
Bowen Pan, Jiankai Sun, Alex Andonian, Aude Oliva, and Bolei Zhou.
\newblock Cross-view semantic segmentation for sensing surroundings.
\newblock {\em arXiv preprint arXiv:1906.03560}, 2019.

\bibitem{roddick2019orthographic}
Thomas Roddick, Alex Kendall, and Roberto Cipolla.
\newblock Orthographic feature transform for monocular 3d object detection.
\newblock {\em Proceedings of the British Machine Vision Conference (BMVC)},
  2019.

\bibitem{samann2018efficient}
Timo S{\"a}mann, Karl Amende, Stefan Milz, Christian Witt, Martin Simon, and
  Johannes Petzold.
\newblock Efficient semantic segmentation for visual bird’s-eye view
  interpretation.
\newblock In {\em International Conference on Intelligent Autonomous Systems},
  pages 679--688. Springer, 2018.

\bibitem{schulter2018learning}
Samuel Schulter, Menghua Zhai, Nathan Jacobs, and Manmohan Chandraker.
\newblock Learning to look around objects for top-view representations of
  outdoor scenes.
\newblock In {\em Proceedings of the European Conference on Computer Vision
  (ECCV)}, pages 787--802, 2018.

\bibitem{thrun2005probabilistic}
Sebastian Thrun, Wolfram Burgard, and Dieter Fox.
\newblock {\em Probabilistic robotics}.
\newblock 2005.

\bibitem{wang2019monocular}
D. {Wang}, C. {Devin}, Q. {Cai}, P. {Krähenbühl}, and T. {Darrell}.
\newblock Monocular plan view networks for autonomous driving.
\newblock In {\em IEEE/RSJ International Conference on Intelligent Robots and
  Systems (IROS)}, pages 2876--2883, 2019.

\bibitem{yang2018hdnet}
Bin Yang, Ming Liang, and Raquel Urtasun.
\newblock Hdnet: Exploiting hd maps for 3d object detection.
\newblock In {\em Conference on Robot Learning}, pages 146--155, 2018.

\bibitem{zhu2018generative}
Xinge Zhu, Zhichao Yin, Jianping Shi, Hongsheng Li, and Dahua Lin.
\newblock Generative adversarial frontal view to bird view synthesis.
\newblock In {\em Proceedings of the IEEE International Conference on 3D Vision
  (3DV)}, pages 454--463, 2018.

\end{thebibliography}
}

\appendix

\begin{figure*}[bt!]

    \centering
    \includegraphics[width=\linewidth]{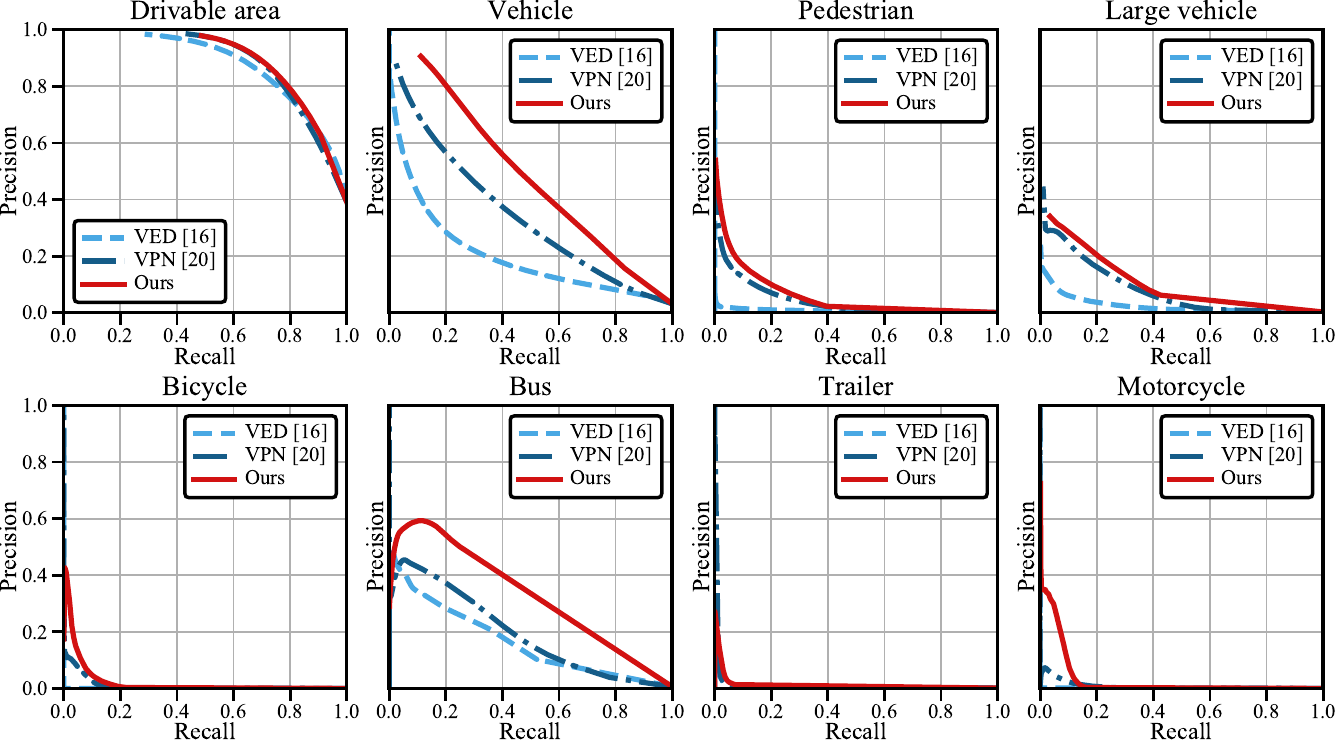}
    \caption{Precision-recall curves for the Argoverse dataset. Best results are curves which occupy the top-right corner of the graph. Our method is more discriminative across all semantic categories, in particular for the \emph{vehicle} and \emph{bus} classes.}
    \label{fig:pr-argoverse}
\end{figure*}

\section{Modifications to competing networks}

In Section 5 of the main paper, we compare our approach to two recent works from the literature: the Variational Encoder-Decoder of Lu\etal\cite{lu2019monocular} and the View Parsing Network of Pan\etal\cite{pan2019cross}. Both works tackle a closely related task to ours, but use other datasets and presume slightly different input dimensions and output map resolutions. In order to compare our work directly, we must therefore make minor architectural changes, which we consider to be the minimum possible for compatibility with our datasets. In the interest of transparency, we detail these changes below:
\begin{description}
\setlength{\itemsep}{0.1em}
    \item[VED:] We modify the bottleneck to use dimensions of
    3\texttimes6\texttimes128 for NuScenes and 4\texttimes7\texttimes128 for Argoverse to account for the different input aspect ratios. We add an additional decoder layer (identical to previous layers) to increase the resolution from 64\texttimes64 to 128\texttimes128 and then bilinearly upsample to our output size of 196\texttimes200.
    \item[VPN:] We increase the transformer module bottleneck dimension to 29\texttimes50 for NuScenes and 38\texttimes50 for Argoverse, and then upsample the output to 196\texttimes 200 using the authors' existing code.
\end{description}
Since we consider a multilabel prediction setting unlike the single label prediction task addressed in the original works, we train both methods using the balanced cross entropy loss described in Section~3.1.

\section{Precision-Recall curves for Argoverse and NuScenes experiments}
\begin{figure*}[hp]
    \centering
    \includegraphics[width=\linewidth]{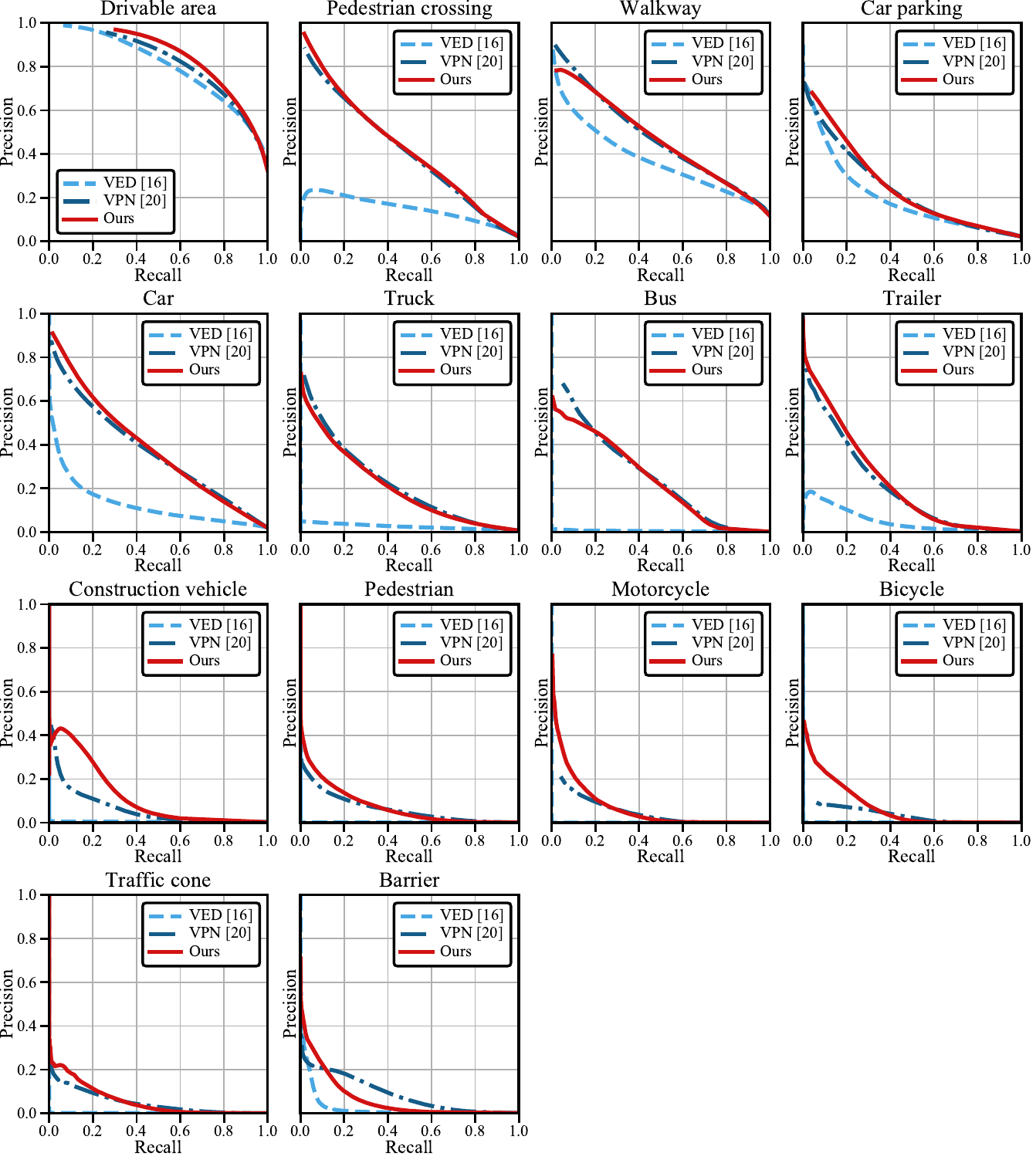}
    \caption{Precision recall curves for the NuScenes dataset.}
    \label{fig:pr-nuscenes}
\end{figure*}

Figures~\ref{fig:pr-argoverse} and~\ref{fig:pr-nuscenes} show the precision-recall trade-off on the Argoverse and NuScenes datasets respectively. Across almost all classes it can be seen that our method represents an upper envelope on the precision achievable for a given recall setting.

\end{document}